\documentclass[letterpaper]{article} 
\usepackage{aaai25}  
\usepackage{times}  
\usepackage{helvet}  
\usepackage{courier}  
\usepackage[hyphens]{url}  
\usepackage{graphicx} 
\usepackage{natbib}  
\usepackage{caption} 
\usepackage{algorithm}
\usepackage{algorithmic}
\usepackage{amsmath}

\usepackage{newfloat}
\usepackage{listings}
\usepackage{booktabs}
\DeclareCaptionStyle{ruled}{labelfont=normalfont,labelsep=colon,strut=off} 
\floatstyle{ruled}
\newfloat{listing}{tb}{lst}{}
\floatname{listing}{Listing}
\title{Multilingual Mathematical Reasoning: Advancing Open-Source LLMs in Hindi and English}
\author{
    Avinash Anand\textsuperscript{\rm 1}, 
    Kritarth Prasad\textsuperscript{\rm 1}, 
    Chhavi Kirtani\textsuperscript{\rm 1}, 
    Ashwin R Nair\textsuperscript{\rm 1}, 
    Manvendra Kumar Nema\textsuperscript{\rm 1}, 
    Raj Jaiswal\textsuperscript{\rm 1}, 
    Rajiv Ratn Shah\textsuperscript{\rm 1}
}
\affiliations{
    \textsuperscript{\rm 1}Indraprastha Institute of Information Technology, Delhi, India\\

     \{avinasha, kritarth20384, chhavi18229, ashwin20037, manvendra22038, Jaiswalp, rajivratn\}@iiitd.ac.in \\
}

\usepackage{bibentry}

\usepackage{fancyhdr}

\fancypagestyle{firstpage}{%
  \fancyhf{} 
  \fancyhead[L]{Pre-print of paper accepted to AAAI 2025}
}

\begin{document}
\thispagestyle{firstpage}
\maketitle

\begin{abstract}
Large Language Models (LLMs) excel in linguistic tasks but struggle with mathematical reasoning, particularly in non-English languages like Hindi. This research aims to enhance the mathematical reasoning skills of smaller, resource-efficient open-source LLMs in both Hindi and English. We evaluate models like OpenHathi 7B, LLaMA-2 7B, WizardMath 7B, Mistral 7B, LLeMMa 7B, MAmmoTH 7B, Gemini Pro, and GPT-4 using zero-shot, few-shot chain-of-thought (CoT) methods, and supervised fine-tuning. Our approach incorporates curriculum learning, progressively training models on increasingly difficult problems, a novel Decomposition Strategy to simplify complex arithmetic operations, and a Structured Solution Design that divides solutions into phases. Our experiments result in notable performance enhancements. WizardMath 7B exceeds Gemini's accuracy on English datasets by +6\% and matches Gemini's performance on Hindi datasets. Adopting a bilingual approach that combines English and Hindi samples achieves results comparable to individual language models, demonstrating the capability to learn mathematical reasoning in both languages. This research highlights the potential for improving mathematical reasoning in open-source LLMs. \footnote{Code, Dataset and Appendix are available at \\ \url{https://github.com/midas-research/Multilingual-Mathematical-Reasoning.git}}
\end{abstract}

\section{Introduction}

Enhancing AI systems to solve complex problems has become a crucial objective within the AI research community, particularly in the realm of mathematical question-answering. While models like GPT-4 and Gemini have demonstrated their strengths in arithmetic \cite{zhang2024careful}, algebra \cite{kao2024solving}, scientific text generation \cite{10.1145/3627673.3679783, anand2023kg}, and symbolic manipulation \cite{dave2024investigating}, they are not without limitations. Our evaluations on the GSM8K \cite{cobbe2021training} and MATH \cite{hendrycks2021measuring} datasets reveal a stark contrast in their capabilities. These models perform well on the relatively straightforward GSM8K dataset, but their effectiveness significantly diminishes when tasked with the more challenging MATH dataset. This dataset includes high-school competition-level questions that require a deeper level of contextual understanding and more advanced reasoning skills. The discrepancies in performance highlight the current limitations of these models in handling complex mathematical problem-solving.


\begin{figure*}[ht]
    \centering
    \includegraphics[width=0.65\textwidth]{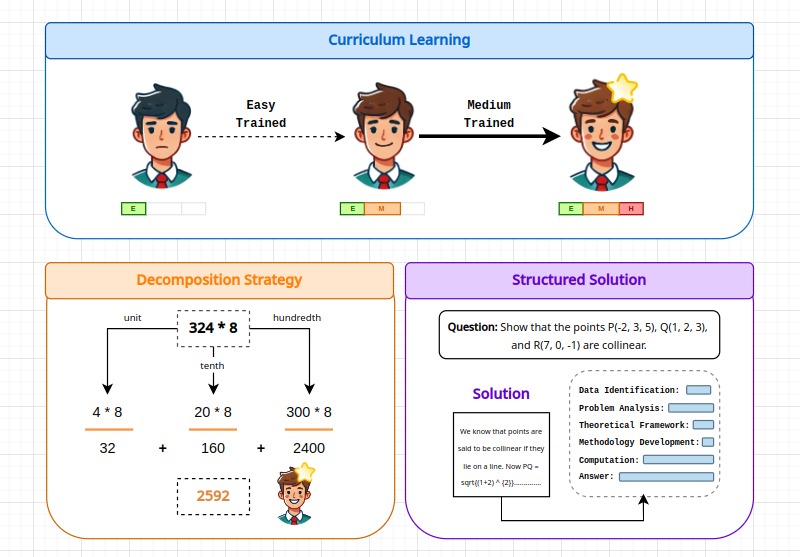}
    \caption{Curriculum Learning with Structured Solutions: A Comprehensive Framework to Gradually Guide Models Through Complex Mathematical Challenges.}
    \label{fig:cl1}
\end{figure*}

In addition to these challenges, there is a noticeable gap in the performance of large language models (LLMs) when applied to English versus non-English languages, particularly in natural language processing tasks such as question answering and classification. This gap is particularly evident in Hindi, India's predominant language, which is used by over 105 million students according to UDISE+ reports for 2019-20\footnote{\url{https://udiseplus.gov.in/udisereport/}}. Enhancing the capabilities of LLMs in Hindi is essential to make these tools more accessible and effective in subject-specific learning contexts. While research efforts such as OpenHathi-7B \cite{sarvam_openhathi}, Hi-NOLIN \cite{nolano_introducing_nolin}, and Airavata \cite{gala2024airavata} have made strides in adapting LLMs to the Hindi language, these models were not originally optimized for domain-specific tasks like mathematical problem-solving.

Recent advancements in open-source LLMs have shown promise in improving mathematical and physics problem-solving abilities \cite{Anand2024GeoVQAAC, anand2024mm, anand2024improving, anand23mathify}, as evidenced by prominent models like WizardMath \cite{luo2023wizardmath}, Mistral \cite{jiang2023mistral}, LLeMMA \cite{azerbayev2023llemma}, and MAmmoTH \cite{yue2023mammoth}. However, these advancements have largely focused on the English language, with limited performance gains observed in Hindi math datasets. Additionally, closed-source models such as GPT-4 and Gemini-Pro continue to outperform open-source models on established benchmarks like GSM8K and MATH, as well as on newly defined Hindi datasets. The significant performance disparity can be attributed to the vast difference in the number of parameters these models are trained on. While the open-source LLMs explored in this research have fewer than 10 billion parameters, well-known closed-source LLMs are trained on considerably large parameter counts. Given the constraints on computational resources, this research focuses on enhancing the performance of smaller open-source LLMs (SLLMs), acknowledging the limitations while seeking to optimize within these parameters.

\begin{figure*}[ht]
    \centering
    \includegraphics[width=0.73\textwidth]{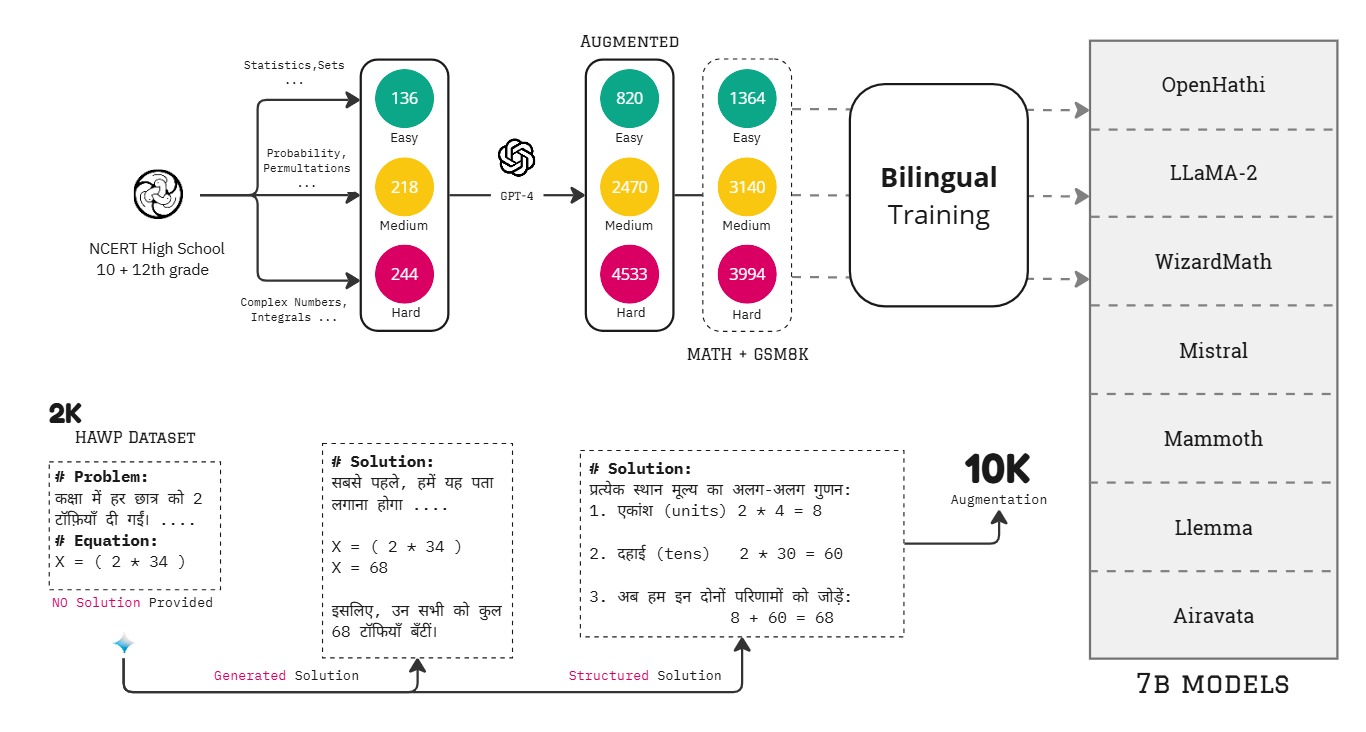}
    \caption{Overall Methodology: The top section illustrates our primary approach, which combines Curriculum Learning and Bilingual Integrated Training. The bottom section depicts the process of applying the decomposition strategy to the HAWP dataset.}
    \label{fig:methodology}
\end{figure*}

This research introduces several key contributions aimed at improving the mathematical capabilities of SLLMs, particularly in Hindi:
\begin{enumerate}
    \item \textbf{Introduction of the Decomposition Strategy:} A novel approach designed to enhance SLLMs' ability to solve complex mathematical operations by breaking them down into smaller, more manageable components in the enhanced HAWP dataset (see Figure \ref{fig:methodology}).
    \item \textbf{Structured Solution Approach with Curriculum Learning:} A combined methodology that integrates a structured solution framework with Curriculum Learning, as illustrated in Figure \ref{fig:cl1}. This approach progressively guides models through increasingly complex mathematical problems, enhancing their problem-solving abilities.
    \item \textbf{Bilingual Combined Training:} We propose the methodology of Bilingual Combined Training, where the Structured Solution Approach with Curriculum Learning is applied on a dataset containing both English and Hindi versions of Mathematical questions-answers.
    \item \textbf{IndiMathQA Dataset Creation:} We developed the IndiMathQA dataset by curating 598 math problems from NCERT \footnote{NCERT is a major textbook for school students in India. \url{https://ncert.nic.in/}} textbooks for grades 10-12, spanning 14 mathematical domains. Expert annotations categorized these into easy, medium, and hard problems. The dataset was expanded to 7,823 questions.
    \item \textbf{Performance Analysis of Multilingual LLMs:} A comprehensive analysis of several LLMs, including five open-source English LLMs, three open-source Hindi LLMs, and two closed-source models, was conducted using both English and Hindi mathematical datasets to assess their capabilities and limitations.
    \item \textbf{Ensuring Transparency and Reproducibility:} To promote transparency and enable future research, the datasets and code developed during this research will be released, with all data having been reviewed by human experts.
\end{enumerate}


\section{Related Work}
Recent advances in large language models (LLMs) have significantly improved their ability to perform complex tasks, particularly in the areas of natural language processing and mathematical reasoning. However, one area that remains underexplored is the application of Curriculum Learning to these models. Originally proposed by Bengio et al. \cite{bengio2009curriculum}, Curriculum Learning is a training strategy that mimics the way humans learn by gradually increasing the complexity of tasks presented to the model. Although widely used in deep learning, its application to LLMs has been limited, particularly in enhancing the models' capabilities in complex, multistep reasoning tasks. This section discusses various open-source and bilingual LLMs, their architectures, and the benchmark datasets used to evaluate their performance.
\subsection{Open-Source Large Language Models}
\begin{enumerate}
\item Llama Model \cite{touvron2023llama}: It is a foundational model requiring less computational power, ideal for fine-tuning for various tasks as it is trained on vast unlabeled data. 
\item Wizard Math \cite{luo2023wizardmath}: It improved mathematical reasoning abilities by applying the "Reinforcement Learning from Evolutionary Instruction Feedback (RLEIF) method" \cite{xu2023wizardlm} in math.
\item MAmmoTH Model \cite{yue2023mammoth}: MAmmoTH was trained using the MathInstruct dataset, a collection of 13 mathematical datasets compiled for fine-tuning.
\item LLeMMA \cite{azerbayev2023llemma}: A model pre-trained on scientific papers, web-based mathematical content and code. It showed advanced mathematical abilities without needing additional fine-tuning.
\item Mistral Model \cite{jiang2023mistral}: A Transformer-based that employs architectural choices to handle complex reasoning tasks in mathematics efficiently without excessive computational overhead.
\end{enumerate}

\subsection{Bilingual Open-Source Large Language Models for English and Hindi}

\subsubsection{OpenHathi-7B}
Sarvam AI used Llama2-7B to create OpenHathi-Hi-v0.1 \cite{sarvam_openhathi}. It integrated a custom tokenizer to expand the embedding layer to 48K tokens and trained on Hindi-English translation and bilingual next-token prediction tasks.

\subsubsection{Hi-NOLIN}
HI-NOLIN \cite{nolano_introducing_nolin}, a Bilingual LLM in the Pythia model suite, was trained on English and code datasets, then further pre-trained on combined Hindi and English data to improve Hindi understanding.
\subsubsection{Airavata}
AIRAVATA model \cite{gala2024airavata} is derived from fine-tuning the OpenHathi model \cite{sarvam_openhathi} with a Hindi instruction-tuned dataset, translated from high-quality English data via IndicTrans2 translation model \cite{gala2023indictrans2}.

\subsection{Benchmark Datasets}

\cite{cobbe2021training} introduced the \textbf{GSM8K} dataset, which consists of 8,500 grade school math problems that require basic arithmetic operations. These problems are designed to be solvable by proficient middle school students. Similarly, \cite{hendrycks2021measuring} released the \textbf{MATH} dataset, containing 12,500 complex problems from high school competitions such as AMC 10 and AMC 12, intended for high school students. It covers topics: Algebra, Counting and Probability, Geometry, Intermediate Algebra, Number Theory, Prealgebra, and Precalculus. \cite{lightman2023let} introduced \textbf{PRM800K}, a dataset with 800,000 step-level feedback labels for solutions to MATH \cite{hendrycks2021measuring} problems, providing annotations ("Positive," "Negative," or "Neutral") to each solution step. For Hindi-speaking students, \cite{sharma-etal-2022-hawp} released \textbf{HAWP} (Hindi Arithmetic Word Problems), which is the only publicly available dataset of Hindi mathematical questions. It is for grades 1 to 6, and comprises 2,336 basic math word problems requiring a single operator solution.


\section{Methodology}

\subsection{Decomposition Strategy on HAWP Dataset}
To improve the computational accuracy of large language models (LLMs) in arithmetic operations involving large numbers, we propose a Decomposition Strategy for multiplication and division tasks. For multiplication, this involves breaking down the multiplicand into place value components—such as hundreds, tens, and ones—and multiplying each by the other multiplicand. The products are then aggregated to obtain the final result. For division, the dividend is similarly decomposed into segments, each divided by the divisor, with the quotients summed to produce the final answer. This has been proposed to combat the poor calculation skills of open-source language models. In this paper, we focus on introducing and validating the Decomposition Strategy using the HAWP dataset, which contains basic mathematical word problems requiring single-operation calculations. This allows us to clearly demonstrate the strategy's effectiveness in a controlled, straightforward context. While exploring its application to more complex datasets is an exciting future direction, we have chosen to concentrate on HAWP for now to ensure a thorough and focused evaluation of this novel approach.


We utilized 2,336 Hindi arithmetic problems from the HAWP dataset, covering basic operations like addition, subtraction, multiplication, and division. Since the original dataset lacked solutions, we enhanced it by generating question-answer pairs using GPT-4, which were then carefully reviewed and corrected by five human experts, resulting in the Enhanced HAWP dataset.

To evaluate the Decomposition Strategy's effectiveness, we applied it to the Enhanced HAWP dataset (see Figure \ref{fig:methodology}). Manually solved examples using this strategy were used in few-shot learning with GPT-4 to modify the remaining solutions in Enhanced HAWP. These refined solutions, with 70\% and 30\% training/testing split, were then used to fine-tune the models OpenHathi 7B, WizardMath-v1.1 7B, and LLeMMa 7B. The resulting accuracy improvements are shown in Table ~\ref{tab:accents}.

In our final phase, we focused on exploring the benefits of fine-tuning using an augmented version of the dataset that we previously prepared Decomposition Strategy-enhanced dataset. We expanded the original 2,000 problems to 10,000 using a one-shot prompting technique with GPT-4. These newly generated samples were carefully reviewed by five human experts for accuracy, resulting in the HMQA (Hindi Math Questions-Answers) dataset. We then used this augmented dataset to fine-tune the models OpenHathi 7B, WizardMath-v1.1 7B, and LLeMMa 7B, with the resulting accuracy compared to previous settings in Table ~\ref{tab:accents}.

\begin{table}[h]
\centering
\begin{tabular}{c|c|c|c|c}

\toprule
\textbf{Model} & \textbf{Add} & \textbf{Sub} & \textbf{Mul} & \textbf{Div} \\
\midrule
\multicolumn{5}{c}{\textit{Zero-shot Prompting}} \\
\midrule
\verb|OpenHathi-7B| & 0.35 & 0.53 & \underline{0.44} & 0.33 \\
\verb|LLaMA-2-7B| & 0.39 & 0.55 & 0.33 & 0.5 \\
\verb|LLeMMA-7B| & 0.49 & 0.63 & 0.22 & 0.17 \\ 
\verb|Mistral-7B| & 0.49 & 0.55 & 0.22 & 0.25 \\ 
\verb|WizardMath-7B|  & \underline{0.63} & \underline{0.67} & 0.22 & \underline{0.67} \\
\verb|Gemini-pro|  & 0.78 & 0.80 & 1.0 & 0.75 \\
\verb|GPT-4|  & 0.98 & 0.93 & 0.88 & 0.91 \\

\midrule
\multicolumn{5}{c}{\textit{Few-shot Prompting}} \\
\midrule
\verb|OpenHathi-7B| & \textbf{0.49} & \textbf{0.63} & 0.34 & \textbf{0.58} \\
\verb|LLaMA-2-7B| & \textbf{0.53} & \textbf{0.72} & \textbf{0.44} & 0.5 \\
\verb|LLeMMA-7B| & \textbf{\underline{0.82}} & \textbf{\underline{0.9}} & \textbf{\underline{1.0}} & \textbf{0.67} \\ 
\verb|Mistal-7B| & \textbf{0.78} & \textbf{0.77} & \textbf{0.56} & \textbf{\underline{0.83}} \\ 
\verb|WizardMath-7B|  & \textbf{0.72} & \textbf{0.73} & \textbf{0.56} & 0.67 \\

\midrule
\multicolumn{5}{c}{\textit{Instruction-Tuning (Enhanced HAWP)}} \\
\midrule
\verb|OpenHathi-7B|  & \textbf{0.78} & \textbf{0.85} & 0.22 & 0.50 \\
\verb|LLeMMA-7B| & 0.78 & 0.83 & 0.67 & 0.67 \\ 
\verb|WizardMath-7B| & \textbf{\underline{0.96}} & \textbf{\underline{1.0}} & \textbf{\underline{0.78}} & \textbf{\underline{0.75}} \\

\midrule
\multicolumn{5}{c}{\textit{Instruction-Tuning (HAWP+Decomposition Strategy)}} \\
\midrule
\verb|OpenHathi-7B| & 0.78 & 0.85 & 0.22 & \textbf{0.67} \\
\verb|LLeMMA-7B|  & \textbf{0.80} & \textbf{0.925} & 0.67 & \textbf{\underline{0.83}} \\ 
\verb|WizardMath-7B|  & \underline{0.95} & \underline{1.0} & \underline{0.78} & \textbf{\underline{0.83}} \\

\midrule
\multicolumn{5}{c}{\textit{Instruction-Tuning (HMQA)}} \\
\midrule
\verb|OpenHathi-7B| & \textbf{0.82} & 0.85 & \textbf{0.44} & \textbf{\underline{0.75}} \\
\verb|LLeMMA-7B|  & \textbf{\underline{0.86}} & \textbf{\underline{0.97}} & \underline{0.67} & \underline{0.75} \\ 
\bottomrule
\end{tabular}
\caption{Performance of LLMs in Hindi Math questions using decomposition strategy. Bold values indicate improvements from the previous step. Underlined values show the highest performance of SLLMs for each operation.}
\label{tab:accents}
\end{table}

\subsection{IndiMathQA}

We have meticulously curated our own comprehensive math problem dataset, referred to as IndiMathQA, sourcing problems from the official NCERT textbooks\footnote{NCERT is a major textbook for school students in India. \url{https://ncert.nic.in/}} used in Indian schools. This dataset contains 598 manually curated math problems and their corresponding solutions. These problems are suited for students in grades 10, 11, and 12, and it encompasses a wide range of problems that vary in complexity and span 14 major mathematical domains, including sets, trigonometry, and the binomial theorem, among others. Appendix provides more details on topic distribution.

\subsection{LLM Enhancement in Bilingual Mathematics}
In this section we demonstrate the strategies used in improving mathematical reasoning skills in Bilingual settings. Our proposed strategies are namely Structured Solution Creation, Curriculum Learning, and Bilingual Training in Hindi and English. We explain the Bilingual Training Dataset Creation in two phases: (i) Classification based on Complexity (required for Curriculum Learning), (ii) Structured Solution Creation and Bilingual Translations. Finally, we demonstrate how we conducted curriculum learning based bilingual fine-tuning on our training datasets.


\subsubsection{Classification based on Complexity}
We have carefully curated a collection of mathematical problems categorized into easy, medium, and hard levels. This collection includes problems from our own dataset as well as from benchmark datasets, such as GSM8K and MATH. Below, we outline the methods we used to classify each problem by its complexity. We utilize additional datasets (GSM8K and MATH) for the sole reason of having more diversity of mathematical topics in our dataset.



\paragraph{\textbf{IndiMathQA}:} The IndiMathQA dataset was carefully annotated by a team of five human experts, resulting in 136 easy, 218 medium, and 244 hard questions. To ensure the reliability of these annotations, we calculated the Average Fleiss' Kappa score, which came out to 0.58, indicating low bias and substantial agreement among the annotators. Further details on the annotation process can be found in the Appendix. This dataset was then augmented to a total of 7823 questions with similar concepts (820 easy, 2,470 medium, and 4,533 hard) using the GPT-4 API, which were then reviewed by a team of 5 human experts to correct any errors to ensure accuracy and consistency (See appendix for augmentation prompt details). 

\vspace{-0.5mm}
 \paragraph{\textbf{GSM8K}:} GSM8K is a grade-school level mathematics dataset, where all questions are generally low in complexity. However, to ensure precision in our classifications, we used LLama 3 (405B) with prompt engineering to assess and rank the questions according to their complexity, based on various criteria, including Language Understanding, Mathematical Complexity, Reasoning Complexity, Number of Variables, and Conceptual Complexity (details in the appendix). For our experiments, we selected the 700 questions with the lowest complexity as the Easy level questions.

 \vspace{-0.7mm}
\paragraph{\textbf{MATH}:} The MATH dataset features competition-level questions for students in grades 8 through 12, with each question annotated by complexity, ranging from Level 1 (easiest) to Level 5 (hardest). For our experiments, we categorized the Level 1 questions in the set as Easy, Levels 2 and 3 as Medium, and the remaining levels as Hard. This classification resulted in 664 Easy, 3,140 Medium, and 3,994 Hard questions. 



\subsubsection{Structured Solution Generation and Language Translations}
LLMs often encounter challenges with hallucinations when solving reasoning tasks. In our manual inspection of base model solutions, we noticed that LLMs sometimes became so focused on solving the problem that they overlooked the underlying theoretical principles required for an accurate solution. This observation aligns with findings from \cite{zheng2023take}, which introduces a novel prompting technique that encourages LLMs to step back and ask questions to better understand the background of a problem. Inspired by this, we hypothesized that by fine-tuning our LLMs on solutions that first pause to consider the theoretical framework, we could guide them to produce more accurate responses. Building on this idea, we didn't stop at just providing the theoretical framework; we went a step further. We designed a comprehensive, step-by-step structured solution format for fine-tuning, which we believe will train the LLM to approach reasoning tasks more methodically and with greater accuracy. To achieve this, we transformed the existing solutions in our datasets into a clear, organized format under the following headings: (i) Data Identification, (ii) Problem Analysis, (iii) Theoretical Framework, (iv) Methodology Development, (v) Computation, and (vi) Answer. (see Figure \ref{fig:cl1})

To guide this process, our team created few-shot examples that illustrate how sample answers should be divided into this structured format. These examples, along with a detailed prompt, were provided to GPT-4, which then generated structured solutions for all problems in our training and testing datasets. A team of 5 human experts then identified and corrected any mistakes in the structured solutions.

Originally, our datasets were in English. After structuring the solutions, we used LLAMA 3 (405B) to translate both the questions and their structured solutions into Hindi. Finally, the English and Hindi versions of GSM8K, MATH, and IndiMathQA are combined on the basis of Easy, Medium, and Hard. This results in a total of 2184 Easy, 5470 Medium, and 8527 Hard problems in our dataset.
\begin{table*}
\centering
\resizebox{\textwidth}{!}{%
\begin{tabular}{l|c|cccccc|cccc}
\toprule
\textbf{Models} & \textbf{Settings} & \multicolumn{6}{c}{\textbf{English Benchmarks}} & \multicolumn{4}{c}{\textbf{Hindi Benchmarks}} \\
\cmidrule(lr){3-8} \cmidrule(lr){9-12}
&  & \textbf{GSM8K} & \textbf{MATH} & \textbf{PRM800K} & \multicolumn{3}{c}{\textbf{EMKB}} & \textbf{Enhanced} & \multicolumn{3}{c}{\textbf{HMKB}} \\
\cmidrule(lr){6-8} \cmidrule(lr){10-12}
&  &  &  &  & \textbf{Easy} & \textbf{Medium} & \textbf{Hard} & \textbf{HAWP} & \textbf{Easy} & \textbf{Medium} & \textbf{Hard}\\
\midrule
LLaMA-7B & \texttt{Base} & 33\% & 22\% & 27\% & 36\% & 28\% & 21\% & 19\% & 17\% & 11\% & 8\% \\
LLeMMA-7B & \texttt{Base} & 14\% & 10\% & 12\%  & 14\% & 12\% & 9\% & 12\% & 11\% & 8\% & 5\% \\
Mistral-7B & \texttt{Base} & 37\% & 23\% & 29\% & 39\% & 30\% & 24\% & 25\% & 22\% & 14\% & 10\% \\
MAmmoTH-7B & \texttt{Base} & 24\% & 14\% & 19\%  & 27\% & 13\% & 11\% & 30\% & 27\% & 22\% & 18\% \\
WizardMath-7B & \texttt{Base} & 71\% & 36\% & 40\% & 64\% & 48\% & 44\% & 68\% & 61\% & 46\% & 36\% \\
\midrule
LLaMA-7B & \texttt{[SFT\_easy]} & 39\% & 25\% & 28\% & 40\% & 29\% & 21\% & 24\% & 22\% & 12\% & 8\% \\
LLeMMA-7B & \texttt{[SFT\_easy]} & 21\% & 11\% & 12\% & 21\% & 13\% & 9\% & 15\% & 14\% & 9\% & 5\% \\
Mistral-7B & \texttt{[SFT\_easy]} & 43\% & 25\% & 31\% & 45\% & 32\% & 24\% & 30\% & 27\% & 15\% & 10\% \\
MAmmoTH-7B & \texttt{[SFT\_easy]} & 30\% & 16\% & 22\% & 33\% & 14\% & 13\% & 41\% & 37\% & 23\% & 18\% \\
WizardMath-7B & \texttt{[SFT\_easy]} & 79\% & 37\% & 42\% & 70\% & 52\% & 44\% & 73\% & 66\% & 47\% & 37\% \\
\midrule
LLaMA-7B & \texttt{[SFT\_easy+medium]} & 42\% & 35\% & 34\% & 41\% & 36\% & 24\% & 25\% & 25\% & 20\% & 16\% \\
LLeMMA-7B & \texttt{[SFT\_easy+medium]} & 21\% & 18\% & 19\% & 21\% & 18\% & 12\% & 15\% & 15\% & 11\% & 10\% \\
Mistral-7B & \texttt{[SFT\_easy+medium]} & 45\% & 37\% & 34\% & 46\% & 39\% & 26\% & 31\% & 29\% & 28\% & 22\% \\
MAmmoTH-7B & \texttt{[SFT\_easy+medium]} & 33\% & 25\% & 30\% & 34\% & 21\% & 15\% & 42\% & 40\% & 32\% & 26\% \\   
WizardMath-7B & \texttt{[SFT\_easy+medium]} & \textbf{80\%} & \textbf{45\%} & \textbf{44\%} & \textbf{73\%} & \textbf{64\%} & \textbf{46\%} & \textbf{77\%} & \textbf{69\%} & \textbf{52\%} & \textbf{42\%} \\
\midrule
\multicolumn{12}{c}{\textit{Bilingual Model Evaluation}} \\
\midrule
OpenHathi-7B & \texttt{Base} & 33\% & 19\% & 24\% & 36\% & 26\% & 20\% & 50\% & 32\% & 28\% & 24\% \\
Airavata-7B & \texttt{Base} & 22\% & 11\% & 15\% & 21\% & 12\% & 9\% & 12\% & 14\% & 10\% & 6\% \\
Hi-NOLIN-9B & \texttt{Base} & 31\% & 16\% & 22\% & 30\% & 21\% & 16\% & 45\% & 30\% & 26\% & 24\% \\
\midrule
\multicolumn{12}{c}{\textit{Closed Source Models}} \\
\midrule
Gemini 1.0 Pro & \texttt{Base} & 75\% & 39\% & 38\% & 68\% & 60\% & 43\% & 81\% & 72\% & 60\% & 48\% \\
GPT-4 & \texttt{Base} & 91\% & 57\% & 70\% & 92\% & 90\% & 81\% & 93\% & 91\% & 83\% & 70\%  \\
\bottomrule
\end{tabular}%
}

\caption{Performance Comparison of Open-Source and Closed-Source Models on English and Hindi Mathematical Benchmarks}
\label{tab:math_benchmarks}
\end{table*}

\subsubsection{Curriculum Learning based Fine-Tuning}
We apply the technique of Curriculum Learning to SLLMs, hypothesizing that by incrementally increasing the complexity of problems during fine-tuning, we can simulate the natural process of human learning—where mastering simpler tasks paves the way for tackling more challenging ones. Our approach utilizes the Easy and Medium datasets, carefully constructed to cover a diverse range of mathematical topics. Each dataset was divided into 70\% for training and 30\% for testing, ensuring this split was consistently applied across all problem categories: easy, medium, and hard.

To implement Curriculum Learning, we first train our SLLMs on the Easy dataset, producing a model checkpoint we refer to as SFT\_Easy. This checkpoint is then further fine-tuned using the Medium dataset, resulting in the final checkpoint, SFT\_Easy+Medium. We evaluate the performance difference between these two checkpoints using testing sets from both benchmark datasets and our curated dataset.

We propose a hypothesis that fine-tuning LLMs on a dataset combining identical question-answer pairs in both English and Hindi could enhance the model's ability to understand and reason through math problems in Hindi—a language where the LLM might not be as proficient. Our reasoning is grounded in the idea that by exposing the LLM to parallel data in English, a language it excels in, the model can leverage its strengths in English to build stronger associations and improve its performance in Hindi.
To test this hypothesis, our Curriculum Learning-based fine-tuning is conducted in two distinct ways: 
\begin{enumerate}
  \item Training the SLLMs separately on English and Hindi datasets, with results presented in Table \ref{tab:math_benchmarks}
  \item Employing Bilingual Combined Training, where the model is trained on a combined dataset of both English and Hindi question-answer pairs. The outcomes of this bilingual training are detailed in Table ~\ref{tab:easy_bilingual} and ~\ref{tab:easy_medium_bilingual}.
  
\end{enumerate}
For our evaluation of SLLMs in Hindi Math reasoning, we only evaluate performance on the Hindi version of IndiMathQA and the HAWP dataset. For the purpose of clarity, we refer the Hindi version of IndiMathQA as HMKB and the English version as EMKB. (Table \ref{tab:math_benchmarks})

\section{Ablation Study}
To comprehensively understand the results and significance of each novel methodology employed in our study, we evaluated performance at every stage. In our experiments with the Hindi dataset, Table~\ref{tab:accents} shows accuracy metrics attained by base models employing zero-shot and few-shot prompting. The findings underscore a substantial performance enhancement with few-shot prompting compared to zero-shot, demonstrating a notable increase of 20-50\% across all operations. This highlights the effectiveness of providing task examples to LLMs. Further, fine-tuning on an enhanced HAWP dataset led to substantial improvements (20-30\% in general) in OpenHathi's performance in addition and subtraction tasks, and in WizardMath's performance across all operations. However, LLeMMA-7B's performance declined after fine-tuning. After manual assessment of its responses, we found that it is exhibiting hallucinations in its solutions. This aligns with recent findings that fine-tuning on new knowledge can increase hallucinations \cite{gekhman2024does}. LLeMMA, primarily pre-trained on English mathematical data, showed hallucinations when provided with new Hindi mathematical knowledge. Our novel Decomposition Strategy significantly enhanced LLeMMA's performance, demonstrating that breaking down complex calculations can reduce hallucinations and enhance reasoning skills. Additionally, addressing hallucinations through augmentation of samples proved effective, as shown by the improvements from instruction-tuning on HMQA for both OpenHathi and LLeMMA. A detailed analysis of the benefits of curriculum learning on both Hindi and English datasets is also provided in the following Results \& Analysis section.

\vspace{-0.6mm}
\section{Result \& Analysis}
In this section, we first examine the impact of Curriculum Learning based fine-tuning in Hindi and English separately. The analysis then explores the results from bilingual combined training. Lastly, we compare the problem-solving capabilities of lightweight open-source models (SLLMs) against closed-source models (LLMs) across different languages and difficulty levels.\vspace{-1.5mm}
\subsection{Curriculum Learning - English Training}
We explore the impact of Curriculum Learning on English Dataset on the performance of SLLMs. In the base setting, the models were fine-tuned on the entire English dataset without distinguishing problem complexity. In this setting, WizardMath-7B demonstrated the highest performance, while LLeMMA-7B exhibited the lowest performance across all benchmarks and our English dataset, EMKB, as shown in Table~\ref{tab:math_benchmarks}. Following this, the models underwent fine-tuning on a subset of easy problems (SFT\_easy), leading to a 4-6\% improvement on easy problems and a 6-8\% increase on the GSM8K benchmark, indicating effective learning of simpler questions during this phase. However, the improvements on more challenging benchmarks like MATH and PRM800K were modest, with only a 1-2\% increase.
In the next stage, models were fine-tuned on both easy and medium problems (SFT\_easy+medium). This approach yielded a consistent 6\% performance increase on medium problems and a 3\% improvement on hard problems. These findings (Table~\ref{tab:math_benchmarks}), suggest that systematically increasing the difficulty of problems enables models to surpass their base setting performance.
\vspace{-0.7mm}
\subsection{Curriculum Learning - Hindi Training}

When applying Curriculum Learning to the Hindi datasets, initially, WizardMath-7B led, while LLeMMA-7B lagged on the Enhanced HAWP Benchmark. Fine-tuning on easy problems (SFT\_easy) improved performance by 3-5\%, but gains on medium and hard problems were minimal. Introducing Curriculum Learning (SFT\_easy+medium) led to an additional 2-4\% improvement on the benchmark and 3-5\% on more difficult problems (Table~\ref{tab:math_benchmarks}). This stepwise training approach effectively enhanced the models' ability to tackle increasingly complex tasks, demonstrating the value of a structured learning regimen in Hindi datasets.
\vspace{-0.7mm}
\subsection{Curriculum Learning - Bilingual Combined Training}
Finally, we tested the performance of SLLMs on full IndiMathQA dataset, covering both Hindi and English versions (Tables ~\ref{tab:easy_bilingual} and ~\ref{tab:easy_medium_bilingual}). SLLMs were fine-tuned using Curriculum Learning on a bilingual combined training set. As a general trend, all models that went through a combined bilingual training (Tables ~\ref{tab:easy_bilingual} and ~\ref{tab:easy_medium_bilingual}) performed better on Hindi Benchmarks in comparison to single language fine-tuning (Table ~\ref{tab:math_benchmarks}). This is a remarkable enhacement achieved from our hypothesis that combined fine-tuning on English and Hindi can help improve model's Hindi Mathematical Reasoning.

Initially, WizardMath-7B achieved the highest performance, while Airavata-7B had the lowest results (Base Settings: Table ~\ref{tab:math_benchmarks}). Fine-tuning on easy problems (SFT\_Easy: Table ~\ref{tab:easy_bilingual}) in both languages led to a consistent 3-5\% improvement on easy questions, enhancing the models' ability to generalize across different linguistic contexts. However, improvements on medium and hard problems were minimal, highlighting the limitations of focusing solely on easy problems.

When fine-tuned on both easy and medium problems in both languages (SFT\_Easy+Medium: Table ~\ref{tab:easy_medium_bilingual}), the models showed more significant gains, with medium problems improving by 11-18\% and hard problems by around 2\%. This demonstrates the effectiveness of Curriculum Learning in enhancing problem-solving abilities and leveraging bilingual training. 
\subsection{Fine-Tuning Open-Source Models (SLLMs)}

In our evaluation of the performance of open-source models such as LLaMA-7B, LLeMMA-7B, Mistral-7B, MAmmoTH-7B, and WizardMath-7B when fine-tuned on combined both Hindi and English versions of IndiMathQA (HMKB and EMKB) (Tables ~\ref{tab:easy_bilingual} and ~\ref{tab:easy_medium_bilingual}), we observed that fine-tuning on both languages combined improves the performance on both the languages significantly compared to the gains when fine-tuning on a single language. As shown in Table ~\ref{tab:easy_bilingual} and ~\ref{tab:easy_medium_bilingual}, fine-tuning on easy problems from both languages led to a marginal 2-3\% performance increase on easy problems in both Hindi and English. This improvement is better than the pre-trained models but less substantial than the improvements seen with single-language fine-tuning, as indicated in Table ~\ref{tab:math_benchmarks}. However, Table ~\ref{tab:easy_bilingual} and ~\ref{tab:easy_medium_bilingual} further demonstrate that fine-tuning easy and medium problems from both languages resulted in a significant major improvement of 11-18\% accuracy.


\subsection{SLLMs (Lightweight open-source) vs LLMs (closed-source)}

WizardMath-7B is the best-performing SLLM in our research. Although GPT-4 performance exceeds even the enhanced performance of WizardMath (Table~\ref{tab:math_benchmarks}, ~\ref{tab:easy_bilingual} and ~\ref{tab:easy_medium_bilingual}), through Curriculum Learning (SFT\_easy+medium) and Bilingual Parallel Training, WizardMath-7B outperforms Gemini 1.0 Pro in English datasets by about 5\% (Table~\ref{tab:math_benchmarks}, ~\ref{tab:easy_bilingual} and ~\ref{tab:easy_medium_bilingual}). This improvement highlights the effectiveness of our methodology in enhancing SLLM's problem-solving abilities in English. However, in Hindi datasets, while WizardMath-7B performance is comparable to Gemini Pro, it still lags by approximately 3\% across Medium and Hard difficulties, likely because WizardMath is more proficient in solving math problems in English than in Hindi.

\subsection{English Models vs Bilingual Models}
Finally, in this comparative analysis of bilingual models and other open-source models (Tables ~\ref{tab:math_benchmarks}, ~\ref{tab:easy_bilingual} and ~\ref{tab:easy_medium_bilingual}), we observe that bilingual models perform consistently better across English and Hindi, unlike most open-source models, except for WizardMath-7B. This consistency is likely due to the language-independent nature of mathematical reasoning. However, bilingual models like OpenHathi-7B, which are not pre-trained on mathematical tasks, show only slight improvement after fine-tuning, suggesting limited learning efficiency. The superior performance of WizardMath-7B highlights the importance of pre-training models on mathematical tasks for robust performance across languages.

\addtolength{\tabcolsep}{-5pt}
\begin{table}[h]
\centering
\begin{tabular}{l@{\hspace{5pt}}|c@{\hspace{5pt}}c@{\hspace{5pt}}c@{\hspace{5pt}}|c@{\hspace{5pt}}c@{\hspace{5pt}}c}
\toprule
\textbf{Models} & \multicolumn{6}{c}{\textbf{IndiMathQA}} \\
\cmidrule(lr){2-7} 
& \multicolumn{3}{c}{\textbf{EMKB}}  & \multicolumn{3}{c}{\textbf{HMKB}} \\
\cmidrule(lr){2-4} \cmidrule(lr){5-7}
& \textbf{Easy} & \textbf{Medium} & \textbf{Hard} & \textbf{Easy} & \textbf{Medium} & \textbf{Hard}\\
\midrule
LLaMA-7B  & \texttt{43\%} & \texttt{31\%} & \texttt{22\%} & \texttt{25\%} & \texttt{16\%} & \texttt{13\%}\\
Llemma-7B  & \texttt{20\%} & \texttt{12\%} & \texttt{9\%} & \texttt{14\%} & \texttt{10\%} & \texttt{7\%} \\
Mistral-7B  & \texttt{48\%} & \texttt{33\%} & \texttt{24\%} & \texttt{30\%} & \texttt{21\%} & \texttt{16\%} \\
Mammoth-7B  & \texttt{36\%} & \texttt{15\%} & \texttt{13\%} & \texttt{40\%} & \texttt{28\%} & \texttt{23\%} \\
WizardMath-7B  & \textbf{\texttt{73\%}} & \textbf{\texttt{64\%}} & \textbf{\texttt{44\%}} & \textbf{\texttt{68\%}} & \textbf{\texttt{46\%}} & \textbf{\texttt{38\%}} \\
\midrule
\multicolumn{7}{c}{\textit{Bilingual Models }} \\
\midrule
OpenHathi-7B  & \texttt{41\%} & \texttt{30\%} & \texttt{23\%} & \texttt{36\%} & \texttt{34\%} & \texttt{26\%} \\
Airavata  & \texttt{23\%} & \texttt{14\%} & \texttt{11\%} & \texttt{16\%} & \texttt{11\%} & \texttt{9\%} \\
Hi-NOLIN  & \texttt{38\%} & \texttt{27\%} & \texttt{25\%} & \texttt{33\%} & \texttt{32\%} & \texttt{25\%} \\
\bottomrule
\end{tabular}%
\caption{Performance of Bilingual Models on IndiMathQA Using SFT\_easy Training}
\label{tab:easy_bilingual}
\end{table}
\addtolength{\tabcolsep}{5pt}

\addtolength{\tabcolsep}{-5pt}
\begin{table}[H]
\centering
\begin{tabular}{l@{\hspace{5pt}}|c@{\hspace{5pt}}c@{\hspace{5pt}}c@{\hspace{5pt}}|c@{\hspace{5pt}}c@{\hspace{5pt}}c}
\toprule
\textbf{Models} & \multicolumn{6}{c}{\textbf{IndiMathQA}} \\
\cmidrule(lr){2-7} 
& \multicolumn{3}{c}{\textbf{EMKB}}  & \multicolumn{3}{c}{\textbf{HMKB}} \\
\cmidrule(lr){2-4} \cmidrule(lr){5-7}
& \textbf{Easy} & \textbf{Medium} & \textbf{Hard} & \textbf{Easy} & \textbf{Medium} & \textbf{Hard}\\
\midrule
LLaMA-7B  & \texttt{44\%} & \texttt{35\%} & \texttt{23\%} & \texttt{29\%} & \texttt{24\%} & \texttt{19\%}\\
Llemma-7B  & \texttt{21\%} & \texttt{14\%} & \texttt{8\%} & \texttt{15\%} & \texttt{10\%} & \texttt{9\%} \\
Mistral-7B  & \texttt{50\%} & \texttt{39\%} & \texttt{29\%} & \texttt{33\%} & \texttt{26\%} & \texttt{22\%} \\
Mammoth-7B  & \texttt{40\%} & \texttt{20\%} & \texttt{18\%} & \texttt{44\%} & \texttt{35\%} & \texttt{27\%} \\
WizardMath-7B  & \textbf{\texttt{75\%}} & \textbf{\texttt{66\%}} & \textbf{\texttt{47\%}} & \textbf{\texttt{72\%}} & \textbf{\texttt{57\%}} & \textbf{\texttt{45\%}} \\
\midrule
\multicolumn{7}{c}{\textit{Bilingual Models }} \\
\midrule
OpenHathi-7B  & \texttt{43\%} & \texttt{33\%} & \texttt{24\%} & \texttt{40\%} & \texttt{37\%} & \texttt{31\%} \\
Airavata  & \texttt{25\%} & \texttt{16\%} & \texttt{13\%} & \texttt{18\%} & \texttt{14\%} & \texttt{11\%} \\
Hi-NOLIN  & \texttt{40\%} & \texttt{30\%} & \texttt{21\%} & \texttt{39\%} & \texttt{35\%} & \texttt{28\%} \\
\bottomrule
\end{tabular}%
\caption{Performance of Bilingual Models on IndiMathQA Using SFT\_easy+medium Training}
\label{tab:easy_medium_bilingual}
\end{table}
\addtolength{\tabcolsep}{5pt}

\section{Conclusion}
This research developed a Bilingual Math Problem Solver using curriculum learning, query decomposition, and structured solution generation. The Decomposition Strategy improved reasoning by breaking down complex queries, Structured Solution addressed the problem of Hallucinations, while curriculum learning enhanced performance on medium and hard problems. WizardMath-7B consistently outperformed other SLLMs (Lightweight open-source) models and often surpassed closed-source models like Gemini 1.0 Pro with these strategies. Our findings demonstrate that integrating these methodologies significantly enhances the problem-solving capabilities of LLMs. Bilingual Parallel Training (Training in multiple languages) provided diverse problem-solving perspectives, proving more effective than single-language training. This study shows how these diverse methodologies can be used to address issues with LLMs in math problem-solving, and can effectively enhance their performance in Hindi.



\section{Acknowledgments}
Dr. Rajiv Ratn Shah is partly supported by the Infosys Center for AI, the Center of Design and New Media, and the Center of Excellence in Healthcare at Indraprastha Institute of Information Technology, Delhi.

\bibliography{aaai25}

\appendix
\section{Appendix}
The following appendix provides a detailed exploration of our research, including: Hyperparameters used and prompts developed, Explanations of key terms in the methodology that were not included in the main paper, Few-shot examples for certain prompts, Additional analysis of the IndiMathQA and MATH datasets (including question types and distribution), A comprehensive error analysis to demonstrate the effect of curriculum learning, and lastly, Annotation analysis of IndiMathQA (including inter-annotator agreements). This supplementary information offers a deeper understanding of our approach and findings.

\section*{Model Parameters and Data Explanation}

\subsection{Model Methodology Overview}

Our research employed an advanced sequence inference technique to fine-tune our model for generating responses to specific prompts. These prompts were meticulously crafted with clear instructions and sample questions to ensure the model produced accurate and varied responses.

\subsubsection{Hyper parameters Used:}
\begin{itemize}
    \item Sampling: Enabled
    \item Top\_k: 40 (for relevance and diversity)
    \item Temperature: 0.8 (for balanced creativity)
    \item Top\_p: 0.90 (for refined nucleus sampling)
    \item Maximum Length: 4096 (for comprehensive answers)
    \item Fine-tuning: 3 epochs
\end{itemize}

\subsection*{Fine-Tuning Prompt}
\textbf{Definition:} This prompt is designed to fine-tune a model by providing a specific instruction and a corresponding response. The focus is on ensuring that the model accurately completes the task while being cautious of incorrect calculations and avoiding repetition of errors.

\begin{verbatim}
Below is an instruction that describes a
task. Write a response that appropriately
completes the request. Be aware of wrong
calculations and do not repeat them.

### Instruction:
{sample['Question']}

### Response:
{sample['Response']}
\end{verbatim}

\subsection*{Data Augmentation Prompt}
\textbf{Definition:} This prompt is used for generating new, conceptually similar questions and answers based on an existing problem. It involves creating problems with varied difficulty levels to enhance the diversity and complexity of the dataset.

\begin{verbatim}
Your task is to create a similar concept-
-ual question and answer with diverse 
difficulty levels (either similarly simp-
-le, the same, or more complex) using 
the provided problem.

Problem:
Question: {sample['Example']}
Answer: {sample['refined_solution']}

New Problem: {sample['Question']}
\end{verbatim}

\subsection*{Decomposition Strategy Prompt}
\textbf{Definition:} This prompt is used for generating updated solutions of HAWP dataset using the Decomposition Strategy, for questions involving multiplication and division. HAWP contains single-operator word problems, so we divided each problem by the type of operation used in the solution. We used separate prompts for Multiplication problems and Division problems. We also gave examples to support the generation.

\begin{verbatim}
Your task is modify the following mathem-
-atical solution by breaking down the mu-
-ltiplicand into place value components
(hundreds, tens, ones, etc.) and then 
multiplying each component by the other
multiplicand. Then, sum the products 
to get the final result. 

Answer: {sample['solution']}

New Answer: {sample['updated_solution']}
\end{verbatim}
One of the examples provided as few-shot example is given below:
\begin{verbatim}
Answer: 543 × 27 = 14661

Updated Answer:
Break down 543 into place value components:

543 = 500 + 40 + 3

Multiply each component by 27:

500 × 27 = 13500

40 × 27 = 1080

3 × 27 = 81

Add the products:
13500 + 1080 + 81 = 14661

Final Answer: 543 multiplied by 27 
equals 14661.
\end{verbatim}

The Decomposition Strategy prompt for division is given below
\begin{verbatim}
Your task is modify the following mathem-
-atical solution by decomposing the 
dividend into segments, then divide 
each by the divisor, and sum the 
quotients to obtain the final answer.

Answer: {sample['solution']}

New Answer: {sample['updated_solution']}
\end{verbatim}
One of the examples provided as few-shot example is given below:
\begin{verbatim}
Answer: 968 ÷ 16 = 60.5

Updated Answer:
Break down 968 into place value 
components:

968 = 800 + 160 + 8

Divide each component by 16:

800 ÷ 16 = 50

160 ÷ 16 = 10

8 ÷ 16 = 0.5
Add the quotients:
50 + 10 + 0.5 = 60.5

Final Answer: 968 divided by 16 
equals 60.5.
\end{verbatim}
\subsection{Explanation of Key Terms and Methodology Components}
In our study, we developed a structured approach to provide solutions for each raw question-answer pair, consisting of the following steps:

\textbf{Data Identification:} 
Specifying the relevant data needed for the problem. This includes identifying variables, constants, and any conditions or parameters related to the problem.

\textbf{Problem Analysis:} 
Examining the problem to understand its components and determine suitable methods for solving it. This includes understanding theoretical aspects such as set theory or integration rules.

\textbf{Theoretical Framework:} 
Establishing the foundational theories and principles that guide the solution. Examples include using set operations for probability problems or antiderivative rules for integrals.

\textbf{Methodology Development:} 
Creating a detailed, step-by-step plan to solve the problem. This includes developing necessary procedures and formulae for the solution.

\textbf{Computation:} 
Performing the calculations and applying the developed methodology to derive the final solution.

\textbf{Solution:}
Presenting the final answer, ensuring it is clear and easy to understand.

\subsection{Data Classification: Easy, Medium, and Hard}

To classify data into easy, medium, and hard categories, we use the following criteria:
\begin{enumerate}
    \item \textbf{Language Understanding:} Assesses the complexity of the language used. Problems with straightforward, clear language are classified as easy, while those with technical or complex phrasing are considered medium or hard.
    
    \item \textbf{Mathematical Complexity:} Assesses the nature of mathematical operations required. Problems involving fundamental concepts are easy, whereas those requiring multiple or intricate operations are medium or hard.

    \item \textbf{Reasoning Complexity:} Evaluates the complexity of reasoning needed. Problems requiring straightforward application of concepts are easy, while those needing detailed logic or multi-step reasoning are medium or hard.

    \item \textbf{Number of Variables:} Evaluates the number of variables involved. Problems with a few variables are easy, whereas those with many or complex variables are medium or hard.

    \item \textbf{Conceptual Complexity:} Assesses the depth of understanding required. Basic conceptual understanding is easy, while advanced or abstract concepts are medium or hard.

\end{enumerate}

\subsection{Structured Problem Solving: Examples of Data Transformation}

We have raw data in the format of \textbf{Question} and \textbf{Final Answer}, which is then converted into a \textbf{Refined Solution}. These examples span various topics, including but not limited to Probability, Linear Algebra, Integration, and many more. Following are examples of few instances from the curated IndiMathQA dataset.
\\

\textbf{Example 1: Conic Sections}

\textbf{Question:} Find the equation of the hyperbola with foci $(0, \pm 3)$ and vertices $\left(0, \pm \frac{\sqrt{11}}{2}\right)$.

\textbf{Raw Solution:}
Given the foci on the y-axis, the equation of the hyperbola is of the form 
\[
\frac{y^{2}}{a^{2}} - \frac{x^{2}}{b^{2}} = 1.
\]

Using the given vertices \(\left(0, \pm \frac{\sqrt{11}}{2}\right)\), we deduce that 
\[
a = \frac{\sqrt{11}}{2}.
\]

The foci are \((0, \pm 3)\), which means 
\[
c = 3.
\]

Using the relationship \(b^{2} = c^{2} - a^{2}\), we find
\[
b^{2} = 3^{2} - \left(\frac{\sqrt{11}}{2}\right)^{2} = 9 - \frac{11}{4} = \frac{36}{4} - \frac{11}{4} = \frac{25}{4}.
\]

This yields the equation of the hyperbola as
\[
\frac{y^{2}}{\left(\frac{11}{4}\right)} - \frac{x^{2}}{\left(\frac{25}{4}\right)} = 1,
\]
which simplifies to
\[
100y^{2} - 44x^{2} = 275.
\]
\\
\textbf{Refined Solution:}

\textbf{Data Identification:}
- Foci: \((0, \pm 3)\)
- Vertices: \(\left(0, \pm \frac{\sqrt{11}}{2}\right)\)

\textbf{Problem Analysis:}
We need to determine the standard form of the equation of the hyperbola given its foci and vertices. For a hyperbola centered at the origin and oriented along the y-axis, its general equation is 
\[
\frac{y^2}{a^2} - \frac{x^2}{b^2} = 1,
\]
where:
- The distance from the center to a vertex is \(a\).
- The distance from the center to a focus is \(c\).
- The relationship \(c^2 = a^2 + b^2\) holds true.
\vspace{+1mm}

\textbf{Theoretical Framework:}
Key concepts and formulae for hyperbolas include:
- Standard form of hyperbola equation: \(\frac{y^2}{a^2} - \frac{x^2}{b^2} = 1\)
- Distance to the vertices: \(\pm a\)
- Distance to the foci: \(\pm c\)
- Relationship: \(c^2 = a^2 + b^2\)

\vspace{+1mm}
\textbf{Methodology Development:}
From the given foci and vertices, we can identify \(a\) and \(c\). Using the relationship \(c^2 = a^2 + b^2\), we can find \(b\) and formulate the hyperbola's equation.
\vspace{+2mm}
\\
\textbf{Computation:}

1. Identify \(c\):
   \[
   c = 3
   \]

2. Identify \(a\):
   \[
   a = \frac{\sqrt{11}}{2}
   \]

3. Apply the relationship \(c^2 = a^2 + b^2\):
   \[
   c^2 = a^2 + b^2 \implies 3^2 = \left(\frac{\sqrt{11}}{2}\right)^2 + b^2
   \]

4. Compute \(a^2\):
   \[
   a^2 = \left(\frac{\sqrt{11}}{2}\right)^2 = \frac{11}{4}
   \]

5. Solve for \(b^2\):
  \[
   9 = \frac{11}{4} + b^2 \implies 9 - \frac{11}{4} = b^2 
   \]
   \[
   \implies \frac{36}{4} - \frac{11}{4} = b^2 \implies \frac{25}{4} = b^2
   \]
   Therefore,
   \[
   b^2 = \frac{25}{4}
   \]
   Therefore,
   \[
   b^2 = \frac{25}{4}
   \]

\textbf{Solution:}
Using the values of \(a^2\) and \(b^2\), the equation of the hyperbola is:
Using the values of \(a^2\) and \(b^2\), the equation of the hyperbola is:
\[
\frac{y^2}{\left(\frac{\sqrt{11}}{2}\right)^2} - \frac{x^2}{\left(\frac{5}{2}\right)^2} = 1 
\]
\[
\implies \frac{y^2}{\frac{11}{4}} - \frac{x^2}{\frac{25}{4}} = 1 
\]
\[
\implies \frac{4y^2}{11} - \frac{4x^2}{25} = 1
\]
Simplifying by multiplying through by 100 to clear the denominators:
\[
100y^2 - 44x^2 = 275
\]

Hence, the equation of the hyperbola is:
\[
\boxed{100y^2 - 44x^2 = 275}
\]\\

\textbf{Example 2: Probability}

\textbf{Question:} A committee of two persons is selected from two men and two women. What is the probability that the committee will have (a) no man? (b) one man? (c) two men?

\textbf{Raw Solution:}
The total number of persons \(=2+2=4\). Out of these four persons, two can be selected in \({ }^{4} \mathrm{C}_{2}\) ways.

(a) No men in the committee of two means there will be two women in the committee.

Out of two women, two can be selected in \({ }^{2} \mathrm{C}_{2}=1\) way.

Therefore, 
\[
\mathrm{P}(\text{no man})=\frac{{ }^{2} \mathrm{C}_{2}}{{ }^{4} \mathrm{C}_{2}}=\frac{1 \times 2 \times 1}{4 \times 3}=\frac{1}{6}
\]

\vspace{5mm}

(b) One man in the committee means that there is one woman. One man out of 2 can be selected in \({ }^{2} \mathrm{C}_{1}\) ways and one woman out of 2 can be selected in \({ }^{2} \mathrm{C}_{1}\) ways. Together they can be selected in \({ }^{2} \mathrm{C}_{1} \times{ }^{2} \mathrm{C}_{1}\) ways.

Therefore, 
\[
\mathrm{P}(\text{One man})=\frac{{ }^{2} \mathrm{C}_{1} \times{ }^{2} \mathrm{C}_{1}}{{ }^{4} \mathrm{C}_{2}}=\frac{2 \times 2}{2 \times 3}=\frac{2}{3}
\]

\vspace{5mm}

(c) Two men can be selected in \({ }^{2} \mathrm{C}_{2}\) way.

Hence,

\[
\mathrm{P}(\text{Two men})=\frac{{ }^{2} \mathrm{C}_{2}}{{ }^{4} \mathrm{C}_{2}}=\frac{1}{{ }^{4} \mathrm{C}_{2}}=\frac{1}{6}
\]

\textbf{Final Answer:}
\begin{itemize}
    \item The probability that the committee will have no man is \(\frac{1}{6}\).
    \item The probability that the committee will have one man is \(\frac{2}{3}\).
    \item The probability that the committee will have two men is \(\frac{1}{6}\).
\end{itemize}

\textbf{Refined Solution:}
\textbf{Data Identification:}
\begin{itemize}
    \item Total people: 2 men and 2 women.
    \item Committee size: 2 persons.
\end{itemize}

\textbf{Problem Analysis:}
\begin{itemize}
    \item Calculate the total number of ways to form a committee of 2 persons from 4 people.
    \item Determine the number of favorable outcomes for:
    \begin{itemize}
        \item No man in the committee.
        \item One man in the committee.
        \item Two men in the committee.
    \end{itemize}
    \item Calculate the probability for each case by dividing the number of favourable outcomes by the total number of ways to form a committee.
\end{itemize}

\textbf{Theoretical Framework:}
\begin{itemize}
    \item Use the combination formula \(C(n, r)\) which denotes the number of ways to choose \(r\) items from \(n\) items without regard to the order, given by \(C(n, r) = \frac{n!}{r!(n-r)!}\).
    \item Probability of an event \(P(E) = \frac{\text{Number of favorable outcomes}}{\text{Total number of outcomes}}\).
    \vspace{+1mm}
\end{itemize}

\vspace{2mm}

\textbf{Methodology Development:}
\begin{enumerate}
    \item Total number of ways to form a committee of 2 out of 4 people: \(C(4, 2)\).
    \item Number of ways to form a committee with:
    \begin{itemize}
        \item No man: \(C(2, 2)\) (choosing 2 women out of 2).
        \item One man: \(C(2, 1) \times C(2, 1)\) (choosing 1 man out of 2 and 1 woman out of 2).
        \item Two men: \(C(2, 2)\) (choosing 2 men out of 2).
    \end{itemize}
\end{enumerate}

\textbf{Computation:}
\begin{enumerate}
    \item Total number of ways to form the committee:
    \[
    C(4, 2) = \frac{4!}{2!(4-2)!} = \frac{4 \times 3}{2 \times 1} = 6
    \]
    \item Number of favorable outcomes:
    \begin{itemize}
        \item No man:
        \[
        C(2, 2) = \frac{2!}{2! \times 0!} = 1
        \]
        \item One man:
        \[
        C(2, 1) \times C(2, 1) = \left(\frac{2!}{1! \times (2-1)!}\right) 
        \]
        \[
        \times \left(\frac{2!}{1! \times (2-1)!}\right) = 2 \times 2 = 4
        \]
        \item Two men:
        \[
        C(2, 2) = \frac{2!}{2! \times 0!} = 1
        \]
    \end{itemize}
    \item Calculate the probabilities:
    \begin{itemize}
        \item No man:
        \[
        P(\text{No man}) = \]
        \[\frac{\text{Number of outcomes for no man}}{\text{Total number of ways}}=\frac{1}{6}
        \]
        \item One man:
        \[
        P(\text{One man}) =\]
        \[\frac{\text{Number of outcomes for one man}}{\text{Total number of ways}}=\frac{4}{6} 
        \]
        \[
        = \frac{2}{3}
        \]
        \item Two men:
        \[
        P(\text{Two men}) = \]
        \[\frac{\text{Number of outcomes for two men}}{\text{Total number of ways}}=\frac{1}{6}
        \]
    \end{itemize}
\end{enumerate}
\textbf{Solution:}
\begin{itemize}
    \item The probability that the committee will have no man is \(\frac{1}{6}\).
    \item The probability that the committee will have one man is \(\frac{2}{3}\).
    \item The probability that the committee will have two men is \(\frac{1}{6}\).
\end{itemize}

These refined solutions provide clear, step-by-step explanations of the problem, ensuring the process is thoroughly documented and easier to follow. This structured approach enhances the accuracy and clarity of solution generation using large language models (LLMs).



\section{Consistency of Topics across Difficulties}
\begin{figure}[ht]
    \centering
    \includegraphics[width=1.1\columnwidth]{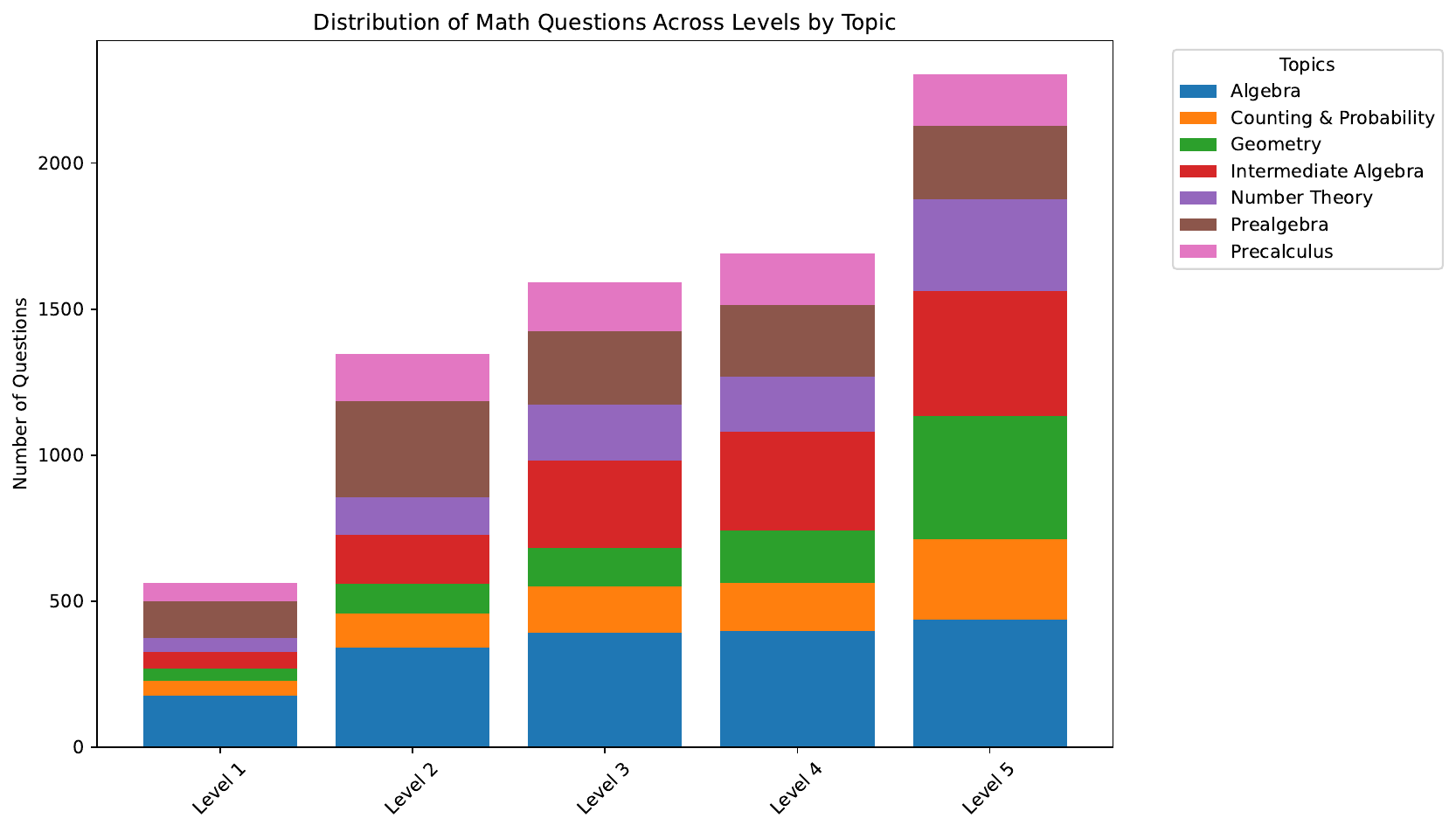}
    \caption{Bar graph showing topic distribution across each difficulty level extracted from the MATH dataset}
    \label{fig:topic-consistency-math}
\end{figure}

\begin{figure}[ht]
    \centering
    \includegraphics[width=1.1\columnwidth]{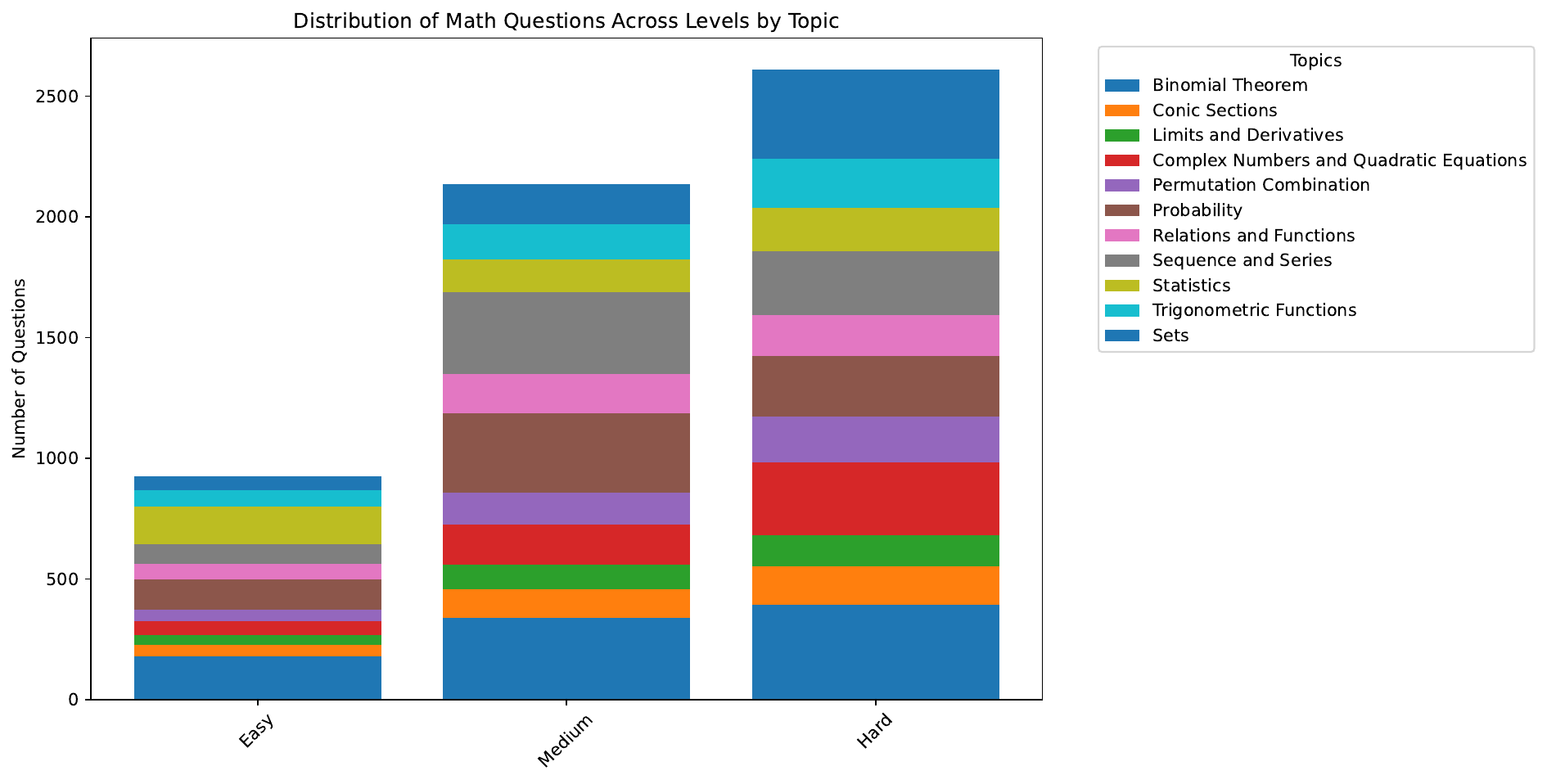}
    \caption{Bar graph showing topic distribution across each difficulty level extracted from the IndiMathQA dataset}
    \label{fig:topic-consistency-IndiMathQA}
\end{figure}

In this section, we demonstrate that during our dataset preparation and categorization of questions into Easy, Medium, and Hard levels, we ensured consistent representation of mathematical topics across all complexity levels in both the MATHS and IndiMath datasets. Figure \ref{fig:topic-consistency-math} and \ref{fig:topic-consistency-IndiMathQA} illustrates the uniform distribution of each mathematical topic at each difficulty level, highlighting the LLM's ability to effectively learn and generalize across topics from easy to hard levels.
remove Math dataset

\section{Curriculum Learning Analysis}
\begin{figure}[ht]
    \centering
    \includegraphics[width=0.8\columnwidth]{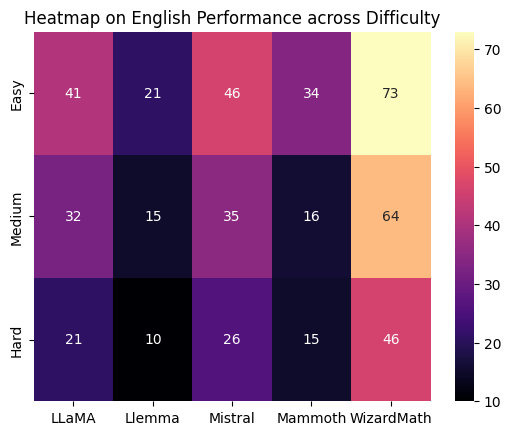}
    \caption{Heatmap showing English mathematical performance across Difficulties (Easy, Medium, Hard) on each Mathematical Models.}
    \label{fig:examples_english}
\end{figure}

\begin{figure}[ht]
    \centering
    \includegraphics[width=0.8\columnwidth]{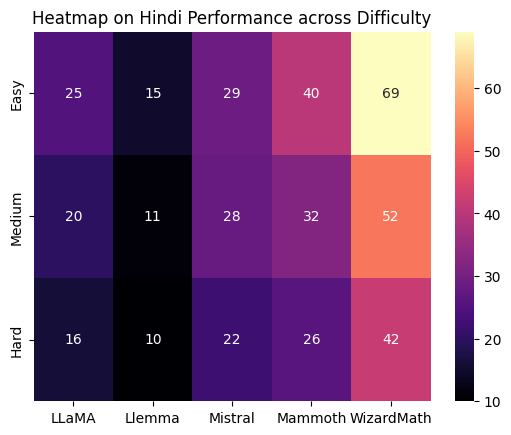}
    \caption{Heatmap showing Hindi mathematical performance across Difficulties (Easy, Medium, Hard) on each Mathematical Model.}
    \label{fig:examples_hindi}
\end{figure}

In this section, we present heatmap visualizations showing each model's final (best) performance for easy, medium, and hard questions in both languages, achieved through our experimentations. Figure \ref{fig:examples_english} and \ref{fig:examples_hindi} show that WizardMath consistently outperforms other open-source counterparts across all difficulties, while LLeMMA performs the worst. These results were obtained after we were able to counter the hallucination problem in LLeMMA to some extent through our strategies as discussed in the Ablation Study (Section 4) of the Main Paper.

\section{Error Analysis: Demonstrating the effect of Curriculum Learning}
Here we present the answers of the WizardMath 7B model before and after SFT based on Curriculum Learning.
\begin{itemize}
    \item \textbf{Question}: A family has two children. What is the probability that both children are boys given that at least one of them is a boy?
    \item \textbf{Wizard Math Base Model Response}: To solve the problem of finding the probability that both children are boys given that at least one of them is a boy, \underline{we assume that if one child is a boy, the other must
    also}
    \underline{be a boy. Therefore, the probability that both children}
    \underline{are boys is 100\%.}
    \item \textbf{Wizard Math SFT\_Easy Response}: To solve the problem of finding the probability that both children are boys given that at least one of them is a boy, we use conditional probability. There are four possible combinations for two children: (Boy, Boy), (Boy, Girl), (Girl, Boy), and (Girl, Girl). Since we know at least one child is a boy, we eliminate the (Girl, Girl) scenario, leaving us with (Boy, Boy), (Boy, Girl), and (Girl, Boy). Each of these scenarios has an equal probability of 1/4. Thus, the probability of at least one child being a boy is 3/4. The probability that both children are boys given this condition is calculated as the probability of both being boys (1/4) divided by the probability of at least one being a boy (3/4), \underline{resulting in 1/2.} Therefore, the probability that both children are boys given that at least one is a boy is 1/2.
    \item \textbf{Wizard Math SFT\_Easy+Medium Response}: To solve the problem of finding the probability that both children are boys given that at least one of them is a boy, we use conditional probability. There are four possible combinations for two children: (Boy, Boy), (Boy, Girl), (Girl, Boy), and (Girl, Girl). Since we know at least one child is a boy, we eliminate the (Girl, Girl) scenario, leaving us with (Boy, Boy), (Boy, Girl), and (Girl, Boy). Each of these scenarios has an equal probability of 1/4. Thus, the probability of at least one child being a boy is 3/4. The probability that both children are boys given this condition is calculated as the probability of both being boys (1/4) divided by the probability of at least one being a boy (3/4), \underline{resulting in 1/3.} Therefore, the probability that both children are boys given that at least one is a boy is 1/3.
    \item \textbf{Comments}: The base model answer demonstrates its inability to apply probability logic correctly. After Curriculum learning with SFT on only Easy examples, the model's reasoning improved substantially, but the model faced an error in calculation at the very last step. This error was rectified when the model was fine-tuned on Easy+Medium, demonstrating the benefit of Curriculum Learning in teaching the LLM the basics of Math, like logical reasoning and calculations, before solving complex questions.
\end{itemize}

\section{Annotation Analysis IndiMathQA }

Our augmented dataset comprises question-answer pairs generated by GPT-4 and verified by five human annotators. This section provides an analysis of the annotation process and the consistency among annotators.

\subsection{Evaluation Methodology and Inter-Annotator Agreement for GPT-4 Generated Responses}

The evaluation methodology for GPT-4 generated responses involved a rigorous review process conducted by five independent human annotators. Each question-answer pair was meticulously evaluated for accuracy, relevance, and clarity. The verification process encompassed all aspects of the data, including data identification, problem analysis, theoretical framework, methodology development, computation, and the final solution. To ensure the reliability of these annotations, inter-annotator agreement was assessed using Fleiss’ Kappa, focusing on two critical aspects.

The first aspect was the validation of solution correctness, where the annotators assessed whether the generated solution appropriately matched the given question. The Fleiss’ Kappa score for this validation was 0.71, indicating substantial agreement among the annotators and confirming the reliability of the model’s output. The second aspect involved classifying the data into easy, medium, and hard categories based on criteria such as language understanding, mathematical complexity, reasoning complexity, number of variables, and conceptual complexity. The Fleiss’ Kappa score for this classification task was 0.58, reflecting moderate agreement. This score suggests that while there was general consensus, some variation existed among the annotators, likely due to subjective interpretations of what constitutes a medium versus a hard problem. Such differences may arise from personal preferences or individual experiences with similar tasks. These Kappa scores collectively validate the robustness of the annotation process, demonstrating a high level of consistency in the human evaluation used to verify and classify GPT-4’s generated responses.

\subsection{Discrepancies and Resolutions}
Discrepancies among the annotators’ evaluations were resolved through a majority voting system. If a majority decision could not be reached due to close scores or ambiguity, a domain expert reviewed the responses to make the final decision. If the domain expert could not reach a consensus, the solution was manually corrected and re-evaluated.
\section{Ethics Statement}
The datasets used for training and testing the LLM were sourced from publicly available repositories. We acknowledge the potential for bias inherent in language model training datasets. The collected data comes from NCERT textbooks, that are used across India for high school education, hence representing the extended demographic within India.

\section{Limitations}
A limitation of our study is that we do not evaluate on all the available sources of data available. Future iterations of this research aim to expand the scope of evaluation, with the help of the research community to collect more comprehensive data from various available sources. Additionally, our study did not include the examination of romanized Hindi sentences, where Hindi word for elephant are written using the English alphabet as "Hathi." This form of input is prevalent in India, particularly when typing on electronic devices. Future research could beneficially extend to enhancing model performance on such inputs as well. The Easy/Medium/Hard distribution of questions was done manually, which can add bias to the distribution as we increase the dataset size.

\end{document}